\documentclass{article} 
\usepackage{iclr2017_conference,times}
\usepackage{hyperref}
\usepackage{url}
\usepackage{graphicx}
\usepackage{booktabs}       
\usepackage{amsfonts}       
\usepackage{amsmath}       
\usepackage{algorithm}
\usepackage{algorithmic}
\usepackage{color}
\usepackage{multirow}
\title{Semi-supervised deep learning by metric embedding}

\author{
  Elad Hoffer\\
  Technion - Israel Institute of Technology\\
  Haifa, Israel\\
  \texttt{ehoffer@tx.technion.ac.il} \\
    \And
  Nir Ailon \thanks{The author acknowledges the generous support of ISF grant number 1271/13}\\
  Technion - Israel Institute of Technology\\
  Haifa, Israel\\
  \texttt{nailon@cs.technion.ac.il}\\
}

\begin{document}

\maketitle

\begin{abstract}
Deep networks are successfully used as classification models yielding state-of-the-art results when trained on a large number of labeled samples. These models, however, are usually much less suited for semi-supervised problems because of their tendency to overfit easily when trained on small amounts of data. In this work we will explore a new training objective that is targeting a semi-supervised regime with only a small subset of labeled data. This criterion is based on a deep metric embedding over distance relations within the set of labeled samples, together with constraints over the embeddings of the unlabeled set. The final learned representations are discriminative in euclidean space, and hence can be used with subsequent nearest-neighbor classification using the labeled samples. 

\end{abstract}

\section{Introduction}
Deep neural networks have been shown to perform very well on various classification problems, often yielding state-of-the-art results.
Key motivation for the use of these models, is the assumption of hierarchical nature of the underlying problem. This assumption is reflected in the structure of NNs, composed of multiple stacked layers
of linear transformations followed by non-linear activation functions. The NN final layer is usually a softmax activated linear transformation indicating the likelihood of each class, which can
be trained by cross-entropy using the known target of each sample, and back-propagated to previous layers. The hierarchical property of NNs has been observed to yield high-quality,
discriminative representations of the input in intermediate layers. These representative features, although not explicitly part of the training objective, were shown to be useful in subsequent tasks in the same domain as demonstrated by \citet{razavian2014cnn}.
One serious problem occurring in neural network is their susceptibility to overfit over the training data. Due to this fact, a considerable part of modern neural network research is devoted to regularization techniques and heuristics such as \citet{dropout, ioffe2015batch, wan2013regularization, szegedy2015rethinking}, to allow the networks to generalize to unseen data samples.
The tendency to overfit is most apparent with problems having a very small number of training examples per class, and these are considered ill-suited to solve with neural network models. Because of this property, semi-supervised regimes in which most data is unlabeled, are considered hard to learn and generalize with NNs.

In this work we will consider a new training criterion designed to be used with deep neural networks in semi-supervised regimes over datasets with a small subset of labeled samples. Instead of a usual cross-entropy between the labeled samples and the ground truth class indicators, we will
use the labeled examples as targets for a metric embedding. Under this embedding, which is the mapping of a parameterized deep network, the features of labeled examples will be grouped together in euclidean space. In addition, we will use these learned embeddings to separate
the unlabeled examples to belong each to a distinct cluster formed by the labeled samples. We will show this constraint translates to a minimum entropy criterion over the embedded distances. Finally, because of the use of euclidean space interpretation of the learned features, we are able to use a subsequent nearest-neighbor classifier to achieve state-of-the-art results on problems with small number of labeled examples.

\section{Related work}
\subsection{Learning metric embedding}

Previous works have shown the possible use of neural networks to learn useful metric embedding. One kind of such metric embedding is the ``Siamese network'' framework introduced by \citet{bromley1993signature} and later used in the works of \citet{chopra2005learning}.
One use for this methods is when the number of classes is  too large or expected to vary over time, as in the case
of face verification, where a face contained in an image has to compared against another image of a face.
This problem was recently tackled by \citet{schroff2015facenet} for training a convolutional network model on triplets of examples. Learning features by metric embedding was also shown by \citet{triplet} to provide competitive classification accuracy compare to conventional cross-entropy regression.
This work is also related to \citet{magnet}, who introduced Magnet loss - a metric embedding approach for fine-grained classification. The Magnet loss  is based on learning the distribution of distances for each sample, from $K$ clusters assigned for each classified class.
It then uses an intermediate k-means clustering, to reposition the different assigned clusters. This proved to allow better accuracy than both margin-based Triplet loss, and softmax regression.
Using metric embedding with neural network was also specifically shown to provide good results in the semi-supervised learning setting as seen in \citet{weston2012deep}.

\subsection{Semi-supervised learning by adversarial regularization}
As stated before, a key approach to generalize from a small training set, is by regularizing the learned model. Regularization techniques can often be interpreted as prior over model parameters or structure, such as $L_p$ regularization over the network weights or activations. More recently,
neural network specific regularizations that induce noise within the training process such as \citet{dropout,wan2013regularization, szegedy2015rethinking} proved to be highly beneficial to avoid overfitting.
Another recent observation by \citet{goodfellow2015explaining} is that training on adversarial examples, inputs that were found to be misclassified under small perturbation, can improve generalization. This fact was explored by \citet{feng2016ensemble} and found to provide notable improvements to the semi supervised regime by \citet{miyato2015distributional}.

\subsection{Semi-supervised learning by auxiliary reconstruction loss}
Recently, a stacked set of denoising auto-encoders architectures showed promising results in both semi-supervised and unsupervised tasks. A stacked what-where autoencoder by \cite{zhao2015stacked} computes a set of complementary variables that enable reconstruction whenever a layer implements a many-to-one mapping. Ladder networks by \cite{rasmus2015semi} - use lateral connections to allow higher levels of an auto-encoder to focus on
invariant abstract features by applying a layer-wise cost function.

Generative adversarial network (GAN) is a recently introduced model that can be used in an unsupervised fashion \cite{goodfellow2014generative}.
Adversarial Generative Models use a set of
networks, one trained to discriminate between data sampled from the true underlying distribution (e.g., a set of images), and a separate
generative network trained to be an adversary trying to confuse the first network.
By propagating the gradient through the paired networks,
the model learns to generate samples that are distributed similarly to the source data. As shown by \cite{radford2015unsupervised}, this model can create useful latent representations for subsequent classification tasks.
The usage for these models for semi-supervised learning was further developed by \citet{springenberg2016iclr} and \citet{salimans2016improved}, by adding a $N+1$ way classifier (number of classes + and additional ``fake'' class) to the discriminator. This proved to
allow excellent accuracy with only a small subset of labeled examples.

\subsection{Semi-supervised learning by entropy minimization}
Another technique for semi-supervised learning introduced by \citet{grandvalet2004semi} is concerned with minimizing the entropy over expected class distribution for unlabeled examples. Regularizing for minimum entropy can be seen as a prior which prefers minimum overlap between observed classes.
This can also be seen as a generalization of the ``self-training'' wrapper method described by \citet{triguero2015self}, in which unlabeled examples are re-introduced after being labeled with the previous classification of the model. This is also related to the ``Transductive suport vector machines'' (TSVM) \citet{tsvm} which introduces a maximum margin objective over both labeled and unlabeled examples.

\section{Our contribution: Neighbor embedding for semi-supervised learning}
In this work we are concerned with a semi-supervised setting, in which learning is done on data of which only a small subset is labeled. Given observed sets of labeled data $X_L=\{(x,y)\}_{i=1}^l$ and unlabeled data $X_U=\{x\}_{i=l+1}^n$ where $x\in \mathcal{X}$, $y\in \mathcal{C}$,
we wish to learn a classifier $f:\mathcal{X} \to \mathcal{C}$ to have a minimum expected error on some unseen test data $X_{test}$.

We will make a couple of assumptions regarding the given data:
\begin{itemize}
\item The number of labeled examples is small compared to the whole observed set $l\ll n$.
\item Structure assumption - samples within the same structure (such as a cluster or manifold) are more likely to share the same label.
This assumption is shared with many other semi-supervised approaches as discussed in \citet{chapelle2009semi},\citet{weston2012deep}.
\end{itemize}
Using these assumptions, we are motivated to learn a metric embedding that forms clusters such that samples can be classified by their $L_2$ distance to
the labeled examples in a nearest-neighbor procedure.

We will now define our learning setting on the semi-labeled data, using a neural network model denoted as $F(x;\theta)$ where x is the input fed into the network, and $\theta$ are the optimized parameters
(dropped henceforward for convenience). The output of the network for each sample is a vector of features of $D$ dimensions $F(x)\in \mathbb{R}^D$ which will be used to represent the input. \\

Our two training objectives which we aim to train our embedding networks by are:
\begin{itemize}
 \item[(i)] Create feature representation that form clusters from the labeled examples $\{(x,y)\}\in X_L$
 such that two examples $x_1,x_2$ sharing the same label $y_1=y_2$ will have a smaller embedded distance than any third example $x_3$ with a different label $y_1\ne y_3$
$$\|F(x_1)-F(x_2)\|_2 < \|F(x_1)-F(x_3)\|_2$$
\item[(ii)]
For each unlabeled example, its feature embedding will be close to the embeddings of one specific label occurring in $L$:

For all $x\in X_U$, $z\in X_L$, there exists a specific class $l\in \mathcal{C}$ such that $$\|F(x)-F(z_l)\|_2\ll \|F(x)-F(z_k)\|_2$$

where $z_l$ is any labeled example of class $l$ and $z_k$ is any example from class $k\in \mathcal{C} \setminus \{l\}$.
\end{itemize}

As the defined objectives create embeddings that target a nearest-neighbor classification with regard to the labeled set, we will refer to it as ``Neighbor embedding''.

\section{Learning by distance comparisons}
We will define a discrete distribution for the embedded distance between a sample $x\in\mathcal{X}$ , and $c$ labeled examples $z_1,...,z_c\in X_L$ each belonging to a different class:
\begin{align}
\label{dist_p}
 P(x;z_1,...,z_c)_i= \frac{e^{-\|F(x)-F(z_i)\|^2}}{\sum_{j=1}^c e^{-\|F(x)-F(z_j)\|^2}} , i \in \{1...c\}
\end{align}
This definition assigns a probability $P(x;z_1,...,z_c)_i$ for sample $x$ to be classified into class $i$, under a 1-nn classification rule, when $z_1,...,z_c$ neighbors are given.
It is similar to the stochastic-nearest-neighbors formulation of \citet{goldberger2004neighbourhood}, and will allow us to state the two underlying objectives as measures over this distribution.
\subsection{Distance ratio criterion}
Addressing objective (i), we will use a sample $x_l\in X_L$ from the labeled set belonging to class $k\in \mathcal{C}$, and another set of sampled labeled examples $z_1,...,z_c\in X_L$. In this work we will sample in uniform over all available samples for each class.

Defining the class-indicator $I(x)$ as
\begin{align*}
\label{onehot}
 I(x_l)_i =
  \begin{cases}
    1       & \quad \text{if } i=k\\
    0  & \quad \text{otherwise}\\
  \end{cases}
\end{align*}

we will minimize the cross-entropy between $I(x_l)$ and the distance-distribution of $x$ with respect to $z_1,...,z_c$\ref{dist_p}:

\begin{align}
 L(x_l, z_1,...,z_c)_L = H\left(I(x_l), P(x_l; z_1,...,z_c) \right)
\end{align}

This is in fact a slightly modified version of distance ratio loss introduced in \cite{triplet}.

\begin{align}
\label{loss_labeled}
L(x_l, z_1,...,z_c)_L=-\log \frac{e^{-\|F(x_l)-F(z_k)\|^2}}{\sum_{i=1}^c e^{-\|F(x_l)-F(z_i)\|^2}}
\end{align}

This loss is aimed to ensure that samples belonging to the same class will be mapped to have a small embedded distance compared to samples from different classes.
\subsection{Minimum entropy criterion}
Another part of the optimized criterion, inspired by \citet{grandvalet2004semi},
is designed to reduce the overlap between the different classes of the unlabeled samples.

We will promote this objective by minimizing the entropy of the underlying distance distribution of $x$, again with respect to labeled samples $z_1,...,z_c$\ref{dist_p}:

\begin{align}
L(x, z_1,...,z_c)_U = H(P(x;z_1,...,z_c))
\end{align}

which is defined as
\begin{align}
\label{loss_unlabeled}
L(x, z_1,...,z_c)_U = -\sum_{i=1}^c  \frac{e^{-\|F(x)-F(z_i)\|^2}}{\sum_{j=1}^c e^{-\|F(x)-F(z_j)\|^2}} \cdot \log{\frac{e^{-\|F(x)-F(z_i)\|^2}}{\sum_{j=1}^c e^{-\|F(x)-F(z_j)\|^2}}}
\end{align}

We note that entropy is lower if the distribution \ref{dist_p} is sparse, and higher if the distribution is dense, and this intuition is compatible with our objectives.

Our final objective will use a sampled set of labeled examples, where each class is represented $\{z_1,...,z_c\}$ and additional labeled $x_l$ and unlabeled $x_u$ examples,
combining a weighted sum of both \ref{loss_labeled} and \ref{loss_unlabeled} to form:

\begin{align}
\label{loss_final}
L(x_l,x_u,\{z_1,...,z_c\}) =  \lambda _L L(x_l, z_1,...,z_c)_L + \lambda_U L(x_u, z_1,...,z_c)_U
\end{align}
Where $\lambda_L, \lambda_U\in [0,1]$ are used to determine the weight assigned to each criterion.

This loss is differentiable and hence can be used for gradient-based training of deep models by existing optimization approaches and back-propagation (\citet{rumelhart1988learning}) through the embedding neural network.
The optimization can further be accelerated computationally by using mini-batches of both labeled and unlabeled examples.
\section{Qualities of neighbor embedding}
We will now discuss some observed properties of neighbor embeddings, and their usefulness to semi-supervised regimes using neural network models.

\subsection{Reducing overfit}
Usually, when using NNs for classification, a cross-entropy loss minimization is employed by using a fixed one-hot indicator (similar to \ref{onehot}) as target for each labeled example, thus maximizing a log-likelihood of the correct label. This form of optimization over a fixed target tend to cause
an overfitting of the neural-network, especially on small labeled sets. This was lately discussed and addressed by \citet{szegedy2015rethinking} using added random noise to the targets,
effectively smoothing the cross-entropy target distribution. This regularization technique was shown empirically to yield
better generalization by reducing the overfitting over the training set. \\
Training on distance ratio comparisons, as shown in our work, provides a natural alternative to this problem. By setting the optimization target to be the embeddings of labeled examples, we create a continuously moving target that is dependent on the current model parameters.
We speculate that this reduces the model's ability to overfit easily on the training data, allowing very small labeled datasets to be exploited.

\subsection{Embedding into euclidean space}
By training the model to create feature embedding that are discriminative with respect to their distance in euclidean space, we can achieve good classification accuracy using a simple nearest-neighbor classifier.
This embedding allows an interpretation of semantic relation in euclidean space, which can be useful for various tasks such as information retrieval, or transfer learning.

\subsection{Incorporating prior knowledge}
We also note that prior knowledge about a problem at hand can be incorporated into the expected measures with respect to the distance distribution \ref{dist_p}.
E.g, knowledge of relative distance between classes can be used to replace $I(x)$ as target distribution in eq. \ref{loss_labeled} and knowledge concerning overlap between classes can be used to relax the constraint in eq. \ref{loss_unlabeled}.

\section{Experiments}
All experiments were conducted using the Torch7 framework by \citet{collobert2011torch7}. Code reproducing these results will by available at \url{https://github.com/eladhoffer/SemiSupContrast}.
For every experiment we chose a small random subset of examples, with a balanced number from each class and denoted by $X_L$.
The remaining training images are used without their labeled to form $X_U$. Finally, we test our final accuracy with a disjoint set of examples $X_{test}$. 
No data augmentation was applied to the training sets.

In each iteration we sampled uniformly a set of labeled examples $z_1,...z_{|\mathcal{C}|}\in X_L$.
In addition, batches of uniformly sampled examples were also sampled again from the labeled set $X_L$, and the unlabeled set $X_U$.

A batch-size of $b=32$ was used for all experiments, totaling a sampled set of $2\cdot b + |\mathcal{C}|$ examples for each iteration, where $|\mathcal{C}|=10$ for both datasets.
We used \ref{loss_final} as optimization criterion, where $\lambda_1=\lambda_2=1$.
Optimization was done using the Accelerated-gradient method by \citet{nesterov1983method}
with an initial learning rate of $lr_0=0.1$ which was decreased by a factor of $10$ after every $30$ epochs. Both datasets were trained on for a total of $90$ epochs. 
Final test accuracy results was achieved by using a k-NN classifier with best results out of $k=\{1,3,5\}$. These results were average over $10$ random subsets of labeled data.

As the embedding model was chosen to be a convolutional network, the spatial properties of input space are crucial.
We thus omit results on permutation-invariant versions of these problems, noting they usually tend to achieve worse classification accuracies.

\subsection{Results on MNIST}
The MNIST database of handwritten digits introduced by \citet{lecun1998gradient} is one of the most studied dataset benchmark
for image classification. The dataset contains 60,000 examples of handwritten digits from 0 to 9 for training and 10,000 additional examples for testing,
where each sample is a 28 x 28 pixel gray level image.

We followed previous works (\citep{weston2012deep},\citep{zhao2015stacked},\citet{rasmus2015semi}) and used semi-supervised regime in which only $100$ samples ($10$ for each class) were used as $X_L$ along with their labels.
For the embedding network, we used a convolutional network with 5-convolutional layers, where each layer is followed by a ReLU non-linearity and batch-normalization layer \citet{ioffe2015batch}.
The full network structure is described in Appendix table \ref{conv_models}.
Results are displayed in table \ref{mnist_conv_results} and reflect that our approach yields state-of-the-art results in this regime.

\begin{figure}[h]
\begin{center}
\includegraphics[trim=70 70 70 70,clip,width=0.7\columnwidth]{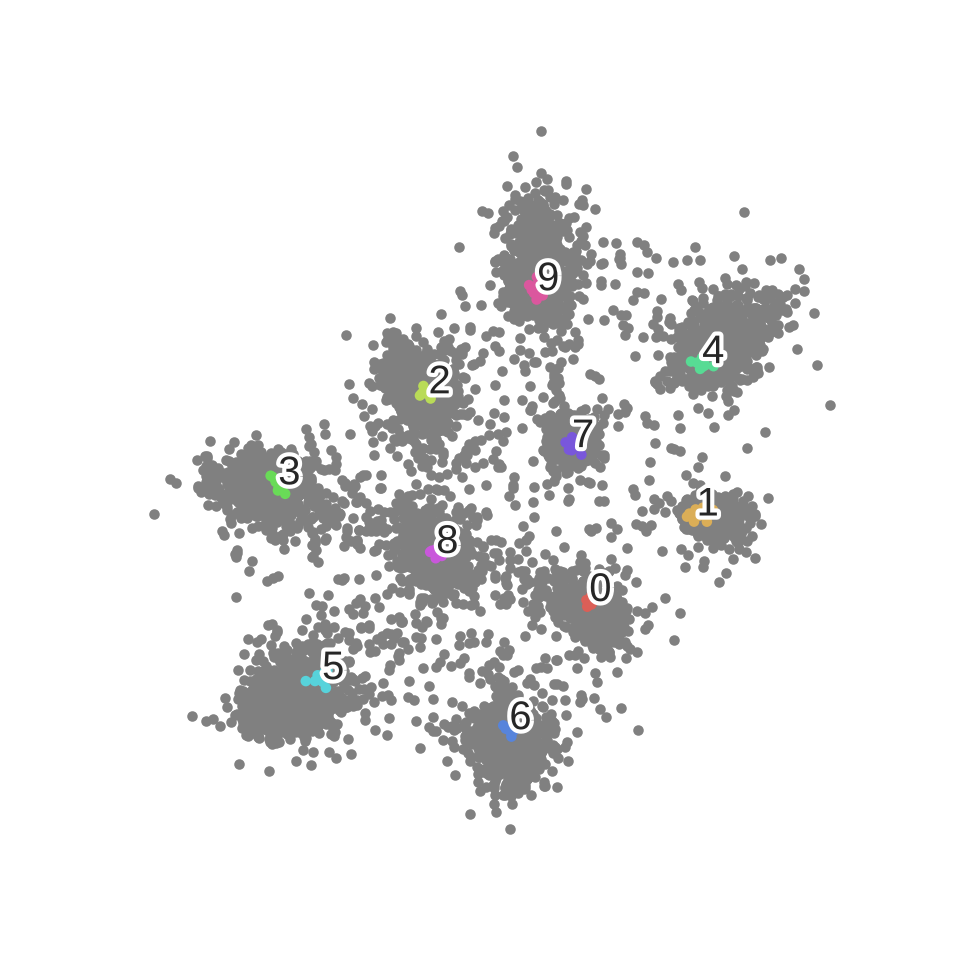}
\end{center}
\caption{MNIST 2d visualization. 100 colored labeled samples, unlabeled samples marked in gray}
\label{mnist_2d}
\end{figure}

We also attempted to visualize the outcome of using this method, by training an additional model with a final 2-dimensional embedding. Figure \ref{mnist_2d} shows the final embeddings, where
labeled examples are marked in color with their respective class, and unlabeled examples are marked in gray.
We can see that, in accordance with our objectives, the labeled examples formed clusters in euclidean space separate by their labels,
while unlabeled examples were largely grouped to belong each to one of these clusters.

\begin{table}[t]
 \caption{Results for MNIST. Using 100 labeled examples, no data-augmentation.
 }
\begin{center}
 \begin{tabular}{llll}
   Model & Test error \% \\
   \hline
   EmbedCNN \citet{weston2012deep} & 7.75   \\
   SWWAE \citet{zhao2015stacked} & 9.17  \\
   Ladder network \citet{rasmus2015semi} & {0.89} ($\pm$ 0.50)   \\
   Conv-CatGAN \citet{springenberg2016iclr}& 1.39 ($\pm$ 0.28) \\
   \hline
   Ours & \bf{0.78} ($\pm$ 0.3)  \\

\end{tabular}
 \end{center}
 \label{mnist_conv_results}
\end{table}

\subsection{Results on Cifar-10}
Cifar-10 introduced by \citet{krizhevsky2009learning} is an image classification benchmark dataset containing $50,000$ training images and $10,000$ test images. The image sizes  $32 \times 32$ pixels, with color.  The classes are
airplanes, automobiles, birds, cats, deer, dogs, frogs, horses, ships and trucks.

Following a commonly used regime, we trained on $4000$ randomly picked samples ($400$ for each class). 
As the convolutional embedding network, we used a network similar to that of \citet{lin2013network} which is described in table \ref{conv_models}.  
The test error results are brought in table \ref{cifar_results}.

As can be observed, we achieve competitive results with state-of-the-art in this regime. 
We also note that current best results are from generative models such as \citet{springenberg2016iclr} and \citet{salimans2016improved} that follow an elaborate and computationally heavy
training procedure compared with our approach.

\begin{table}[t]
 \caption{Results for Cifar-10. Using 4000 labeled samples, no data-augmentation.}
\begin{center}
 \begin{tabular}{ll}
   Model & Test error  \%\\
   \hline
   Spike-and-Slab Sparse Coding \citet{goodfellow2012large} & 31.9  \\
   View-Invariant k-means \cite{hui2013direct} & 27.4 ($\pm$ 0.7)  \\
   Exemplar-CNN \citet{dosovitskiy2014discriminative} & 23.4 ($\pm$ 0.2) \\
   Ladder network \citet{rasmus2015semi} & 20.04 ($\pm$ 0.47) \\
   Conv-CatGan \citet{springenberg2016iclr} & 19.58 ($\pm$ 0.58) \\
   ImprovedGan \citet{salimans2016improved} & \bf{18.63 ($\pm$ 2.32)} \\
   \hline
   Ours & 20.3 ($\pm$ 0.5)  \\

\end{tabular}
 \end{center}
 \label{cifar_results}
\end{table}
\section{Conclusions}
In this work we have shown how neural networks can be used to learn in a semi-supervised setting using small sets of labeled data, by replacing the classification objective with a metric embedding one.
We introduced an objective for semi-supervised learning formulated as minimization of entropy over a distance encoding distribution. This objective is compliant with standard techniques of training
deep neural network and requires no modification of the embedding model.
Using the method in this work, we were able to achieve state-of-the-art results on MNIST with only $100$ labeled examples and competitive results on Cifar10 dataset. We speculate that this form of learning
is beneficial to neural network models by decreasing their tendency to overfit over small sets of training data.
The objectives formulated here can potentially leverage prior knowledge on the distribution of classes or samples, as well as incorporating this knowledge in the training process.
For example, utilizing the learned embedded distance, we speculate that a better sampling can be done instead of a uniform one over the entire set.

Further exploration is needed to apply this method to large scale problems, spanning a large number of available classes, which we leave to future work.

\bibliography{semisup}
\bibliographystyle{iclr2017_conference}

\section{Appendix}

\begin{table}[ht]
\caption{Convolutional models - (feature-maps, kernel, stride) for each layer. Convolutional layers are each followed by ReLU and Batch-norm.}
\label{conv_models}
\begin{center}
\begin{tabular}{l|l}
\multicolumn{2}{c}{\bf Model} \\
\hline
MNIST         &   Cifar-10  \\
\hline
Input:  $28 \times 28$ monochrome& Input:  $32 \times 32$ RGB\\
\hline
Conv-ReLU-BN (16, 5x5, 1x1)  & Conv-ReLU-BN (192, 5x5, 1x1) \\
Max-Pooling (2x2, 2x2)       & Conv-ReLU-BN (160, 1x1, 1x1) \\
Conv-ReLU-BN (32, 3x3, 1x1)  & Conv-ReLU-BN (96, 1x1, 1x1) \\
Conv-ReLU-BN (64, 3x3, 1x1)  & Max-Pooling (3x3, 2x2) \\
Conv-ReLU-BN (64, 3x3, 1x1)  & Conv-ReLU-BN (96, 5x5, 1x1) \\
Max-Pooling (2x2, 2x2)       & Conv-ReLU-BN (192, 1x1, 1x1) \\
Conv-ReLU-BN (128, 3x3, 1x1) & Conv-ReLU-BN (192, 1x1, 1x1) \\
Avg-Pooling (6x6, 1x1)       & Max-Pooling (3x3, 2x2) \\
                             & Conv-ReLU-BN (192, 3x3, 1x1) \\
                             & Conv-ReLU-BN (192, 1x1, 1x1) \\
                             & Avg-Pooling (7x7, 1x1) \\
\hline
\end{tabular}
\end{center}
\end{table}

\end{document}